\documentclass[a4paper]{article}

\usepackage{INTERSPEECH2021}
\usepackage{times}
\usepackage{latexsym}
\usepackage[T1]{fontenc}
\usepackage{url,graphicx}
\usepackage{verbatim}
\usepackage{booktabs} 
\usepackage{caption}
\usepackage{xcolor}
\usepackage{pifont}
\usepackage{subcaption}
\usepackage{enumitem}
\usepackage{hyperref}
\graphicspath{ {fig/} }

\title{Adapting Long Context NLM for ASR Rescoring in Conversational Agents}
\name{Ashish Shenoy, Sravan Bodapati, Monica Sunkara, Srikanth Ronanki, Katrin Kirchhoff}
\address{
  Amazon AWS AI, USA}
\email{\{ashenoy, sravanb, sunkaral, ronanks, katrinki\}@amazon.com}

\begin{document}

\maketitle
\begin{abstract}
Neural Language Models (NLM), when trained and evaluated with context spanning multiple utterances, have been shown to consistently outperform both conventional n-gram language models and NLMs that use limited context. In this paper, we investigate various techniques to incorporate turn based context history into both recurrent (LSTM) and Transformer-XL based NLMs. For recurrent based NLMs, we explore context carry over mechanism and feature based augmentation, where we incorporate other forms of contextual information such as bot response and system dialogue acts as classified by a Natural Language Understanding (NLU) model. To mitigate the sharp nearby, fuzzy far away problem with contextual NLM, we propose the use of attention layer over lexical metadata to improve feature based augmentation. Additionally, we adapt our contextual NLM towards user provided on-the-fly speech patterns by leveraging encodings from a large pre-trained masked language model and performing fusion with a Transformer-XL based NLM. We test our proposed models using N-best rescoring of ASR hypotheses of task-oriented dialogues and also evaluate on downstream NLU tasks such as intent classification and slot labeling.  The best performing model shows a relative WER between 1.6\% and 9.1\% and a slot labeling F1 score improvement of 4\% over non-contextual baselines.

\end{abstract}

\section{Introduction}

Conversations in goal or task oriented conversational interfaces, such as digital personal assistants or chatbots, typically span multiple turns of back and forth between a user and a bot \cite{multiwoz:18,houseini_2020}. These interactions by definition involve accomplishing a specific task in a particular domain and mostly adhere to a dialogue structure that can be determined beforehand. Hence, most real-time task oriented chatbot systems allow users to provide a dialogue grammar that tries to capture usage patterns, intents, slots, and the conversation structure of the interactions \cite{petar:2015,lemon-etal-2006-isu, aggandhe:18}.

A conventional NLM rescorer for ASR is typically trained and evaluated on context that is limited to single turns \cite{mikolov:15,bengio:01} and is therefore sub-optimal in task oriented dialogue systems. A number of studies recently explored incorporating cross utterance context into both recurrent and non-recurrent NLMs. In particular, recurrent neural network (RNN) and LSTM based NLMs are trained and evaluated without resetting hidden states across sentences \cite{ms:2018, irie:19, xiong-etal-2018-session}. The work proposed in \cite{partha:19} uses a multi-head dot-product attention over LSTM hidden states to better exploit the contextual information. However, context carry over models with either RNNs or LSTMs each has its own limitations. The wide spread adoption of transformer architecture based on self attention \cite{vaswani:17} and pretrained masked language models \cite{devlin:18} paved the way for use of transformers for NLMs. In transformer NLMs, long span context modeling is achieved through longer or adaptive attention spans \cite{sukhbaatar-etal-2019-adaptive}. While the initial work used modified attention masks to handle longer input sequences more efficiently, Transformer-XL (TXL) \cite{dai-etal-2019-transformer} made use of segment wise recurrence, making it well suited for long span decoding. More recently, \cite{sun2021transformer} further tried to improve cross utterance decoding using TXL by adding a LSTM fusion layer, where the hidden states of LSTM are carried over multiple sentences. The work in \cite{kim-etal-2019-gated} explored the use of a conversational context embedding from recent history to improve the contextualization of end-to-end ASR models where as some other works \cite{mikolov:12,keli:2018,anirudh:18,Chen:2015} used an explicit topic vector or a neural cache and a domain classifier for domain and contextual adaptation. 

While vast majority of the previous work focused on using context spanning multiple utterances, they do not take into account other contextual signals from NLU such as dialogue act, predefined dialogue grammars or any user-provided speech patterns. 
In this work, we investigate a number of ways to improve contextualization of LSTM and TXL based NLMs to transcribe task-oriented dialogue audio. Specifically, we use long span context that spans across all the utterances in the same dialogue session and system dialogue acts as classified by a NLU model. Additionally, in order to adapt the NLM towards user provided speech patterns in the pre-defined dialogue grammar, we use semantic embeddings derived from a large pretrained masked language models such as BERT \cite{devlin:18}. We use perplexity (PPL) and word error rate (WER) as our ASR evaluation metrics and also evaluate on the natural language understanding (NLU) metrics such as intent classification (IC) and slot labeling (SL) F1 scores to measure the impact on end to end task success rate. The overall contributions of this work can be summarized as follows :
\begin{itemize}[topsep=0pt, leftmargin=*]
    \setlength{\itemsep}{0pt}
    \setlength{\parskip}{0pt}
    \setlength{\parsep}{0pt}
    \item For speech recognition in task-oriented conversations, we show that utilizing long span context from past utterances in the same dialogue session along with system dialogue act, provides significant improvements in WER reduction (WERR).
    \item We propose a new architecture that lets us leverage user provided speech patterns by using embeddings derived from a pretrained masked language model, such as BERT, to perform on-the-fly adaptation of a neural rescorer.
    \item By combining the different forms of contextual information, we successfully train NLMs that achieve a WERR ranging from 1.6\% to 9.1\%, IC F1 improvement ranging from 0.3\% to 1.2\% and SL F1 ranging from 0.4\% to 4.5\% over a non-contextual LSTM LM baseline.
\end{itemize}

\section{Approach}

\subsection{Recurrent neural language models}

\subsubsection{Baseline LSTM-based NLM}
A typical language model in an ASR system computes the probability of a sequence of words $W = w_0,...,w_N$ auto-regressively as:
\begin{equation}\label{eq:1}
p(W) =\prod_{i=1}^N{p(w_i|w_{<i})}
\end{equation}
We use a standard LSTM-based neural network that can model the probability of a word given its history as described in equation \ref{eq:1}. If ${w_0, w_1..,w_n}$ be a sequence of words in a dialogue turn, the probability of $p(w_i|w_{<i})$ can be summarized as below 
\begin{equation}\label{eq:3}
\begin{aligned}
&embed_i = E^T_{ke}w_{i-1} \\
& c_i, h_i = LSTM(h_{i-1}, c_{i-1}, embed_i) \\
&p(w_i|w_{<i}) = Softmax(W^T_{ho}h_i) \\
\end{aligned}
\end{equation}
where $embed_i$ is a fixed size lower dimensional word embedding, the output from the LSTM is projected to word level outputs using $W^T_{ho}$ and then a $Softmax$ layer converts the word level outputs into final word level probabilities.

\subsubsection{Long-span LSTM-based NLM with context carry over}
One way of achieving contextual language modeling in LSTM-LMs is to simply carry over the fixed size context vectors $h_n$ and $c_n$ after the last time step in the previous turn to the initial state in the current turn. For this purpose, we use (a) system dialogue act, and (b) previous bot response to prime the model for the current user turn. The dialogue act (DA) and the bot response (BR) are concatenated and separated with an explicit tag (such as "<dialog\_act>" and "<bot\_response>") that we define and include in the vocabulary. The gradient is backpropogated only for user turns.

\subsubsection{Feature augmentation using context embeddings}
It is known that LSTM-based NLMs with context carry over (CCO) suffer from sharp nearby, fuzzy far away issues \cite{khandelwal-etal-2018-sharp} and therefore we propose to investigate input feature augmentation using various context embeddings. \\ 

\noindent \textbf{Average word embedding}: A standard word embedding module produces word-wise embedding vectors over contextual representations; the final fixed length embedding is obtained by averaging embedding vectors over the entire contextual sequence. 
\begin{equation}
\begin{aligned}
&embed_{ctx} = \langle E^T_{ke}[ctx_1;ctx_2...;ctx_k] \rangle \\
&\hat{embed_i} = [embed_{ctx};embed_i]
\end{aligned}
\vspace{-2mm}
\end{equation}
where $embed_i$ is the input word embedding at time step $i$.\\ 

\noindent \textbf{Attention word embedding}: Some parts of the contextual information may be more important than the others. To allow for this we add an explicit attention layer over the context. The context representation embeddings are passed through a weighted attention layer. The output embedding from the weighted attention layer is concatenated with the input word embedding. The attention layer can be summarized as follows and is similar to \cite{bahdanau:16} and \cite{yang-etal-2016-hierarchical}. Here $e_{t}$ is the embedding for $t$-th word in context and $u_w$ is the query word embedding in the utterance being decoded:
\begin{equation}
\begin{aligned}
&u_{t} = tanh(W_{w}e_{t}+b_w) \\
&a_{t} = \frac{exp(u_{t}u_w)^\top}{\sum _{t}exp(u_{t}u_w)^\top} \\
&s_i = \sum_{t}a_{t}h_{t} \\
&\hat{embed_i} = [s_i;embed_i] \\
\end{aligned}
\end{equation}

\noindent \textbf{Semantic embedding}: Pretrained masked language models (MLM) trained on large scale unlabelled data, are capable of learning richer semantic language representations and are powerful language learners \cite{Radford2019,brown2020language}. Specifically, at training time, we use BERT to obtain embeddings from the CLS token of all the training sentences separated by domain. The CLS token is a special classification token that is prefixed to every sequence and the final hidden state corresponding to this token is used as the aggregate sequence representation for classification tasks \cite{devlin:18, sun:19}. Then for every domain, we average these embeddings to obtain a single fixed length domain embedding.
At inference time, we use these user provided speech patterns or sample utterances to obtain a domain representation for the bot. We then use cosine similarity to pick the closest domain embedding seen at training time to adapt the NLM towards these speech patterns.

\subsection{Non recurrent neural language models}

\begin{figure}[!t]
    \includegraphics[width=0.8\linewidth]{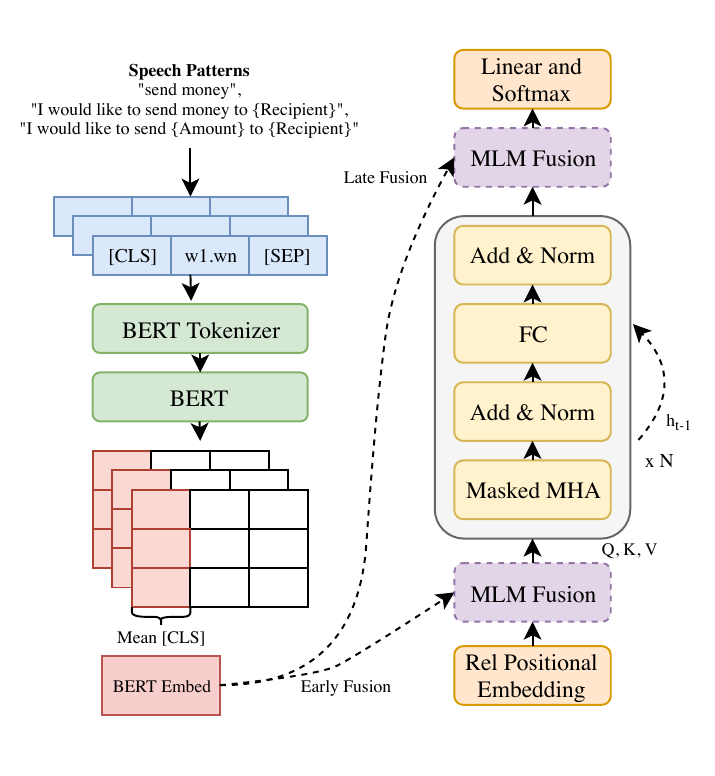}
    \caption{Transformer-XL model architecture showing the early and late MLM fusion layer.}
    \label{fig:txl_bert_fusion}
    \vspace{-6mm}
\end{figure}

\subsubsection{Transformer-XL based NLM}
To address the limitations of using a fixed-length context, Transformer-XL \cite{dai-etal-2019-transformer} adds a recurrence mechanism to the Transformer architecture. Hence, we directly experiment with the TXL models where we carry over context across sentences. During training, the hidden state sequence computed for the previous segment is reused while processing a new segment as an extended context. The final model consists of a stack of decoder blocks, where each block includes a multi-headed attention (MHA) layer with a residual connection and a set of full-connected (FC) layers followed finally by a linear and softmax layer. The MHA layer uses upper triangular masks in order to ensure uni-directionality while decoding. Formally, if $s_t = [x_{t,1}, ..., x_{t,L}]$ and $s_{t+1} = [x_{t+1,1},...,x_{t+1,L}]$ are two consecutive segments of length $L$ and $h^n_t$ is the $n$-th layer hidden state produced for the $t$-th segment $s_t$, then, the $n$-th layer hidden state for segment $s_{t+1}$ is produced as follows:
\begin{equation}
\begin{aligned}
\tilde{h}^{n-1}_{t+1} = [SG(h_t^{n-1}) \circ h_{t+1}^{n-1}] \\
q^n_{t+1}, k^n_{t+1}, v^n_{t+1} = \tilde{h}_{t+1}^{n-1}{W_q}^\intercal \\
h^n_{t+1} = TransformerLayer(q_{t+1}^n, k^n_{t+1}, v^n_{t+1})
\end{aligned}
\end{equation}
where $SG(.)$ stands for stop gradient.

\begin{table}[t]
\footnotesize
\centering
\begin{tabular}{l|c|c}
\toprule
\textbf{Source} & \textbf{\# Dialogues} & \textbf{\# Utterances}\\
\midrule
DSTC8 SGD & 22825 & 231642 \\
MultiWOZ 2.1 & 10420 & 143048 \\
MultiDoGo & 102870 & 1376816 \\
\bottomrule
\end{tabular}
\caption{Goal-oriented datasets and sizes used}
\label{tab:datasets}
\vspace{-6mm}
\end{table}

\subsubsection{MLM fusion with Transformer-XL}
We explore several different techniques of fusing the semantic embeddings derived from pretrained masked language models (MLMs) with TXL. They can be broadly categorized as early and late fusion as shown in Figure \ref{fig:txl_bert_fusion}. In late fusion we explore two different methods : Simple and Cold fusion similar to \cite{bert_fusion:20} and \cite{sunkaral:2020}. \\ 
\textbf{Early Fusion} : In this technique, we explore input level fusion by combining the input word embedding with BERT derived embedding at every time step. The computation procedure can be summarized as:
\begin{equation}\label{eq:6}
g_t = \sigma(W[E_t; e^{MLM}] + b)
\vspace{-2mm}
\end{equation} 
\\
\textbf{Simple Fusion} : In this form of fusion the hidden state from the last layer of the TXL decoder is concatenated with the BERT derived embedding and is followed by a single projection layer with a non-linear activation function $\sigma$, such as $sigmoid$. \\
\begin{equation}\label{eq:6}
g_t = \sigma(W[h_t^{TXL}; e^{MLM}] + b)
\vspace{-2mm}
\end{equation} \\
\textbf{Cold Fusion} :
This is a more complex fusion that is achieved by introducing gates. The gated fusion allows moderation of information flow between the TXL and the BERT derived embedding during training. Our approach is a modified cold fusion approach from \cite{sriram:2017} which is as follows: \\
\begin{equation}
\begin{aligned}
\widehat{e^{MLM}} = \sigma (W[e^{MLM}] + b)\\
g_t = \sigma (W[\widehat{e^{MLM}}; h_{t}^{TXL}]+b)\\
h_t^{CF} = [h_t^{TXL}; (g_t \circ e^{MLM})]\\
r_t^{CF} = \sigma (W[h_t^{CF}] + b)
\vspace{-2mm}
\end{aligned}
\end{equation} 

Where $E_t$ is input word embedding, $h_t^{TXL}$ is the hidden state from the last transformer decoder, $e^{MLM}$ is the BERT derived embedding from in domain sample utterances and $\circ$ stands for hadamard product. In all the above methods, we pass the output from the fusion layer to a linear layer followed by softmax to predict the next word in the sequence. \\

\section{Experimental Setup}
\subsection{Dataset}
We use a combination of task-oriented user-bot dialogues along with actor, domain and dialogue act annotated for LM training: Schema-Guided Dialogue Dataset \cite{Rastogi:20}, MultiWOZ 2.1 \cite{multiwoz:18} \cite{MultiWoz:19} and a small sample of MultiDoGo \cite{multidogo:19}. These are user-agent conversations with an average of 10 turns per dialogue. The average length of a sequence in a single dialogue session where all turns are concatenated was around 120. The statistics of the data sets are listed in Table \ref{tab:datasets}. The merged text only dataset consisted of 260,415 training samples, 51,602 validation samples, 56,091 test samples and around 9.9 million running words. The final vocabulary contained 25,000 most frequent words, which included words from bot responses and user samples. The out-of-vocabulary words were modeled with the \verb|<unk>| token. Since we used a combination of data sources annotated with different tagging schemes, we had to normalize the dialogue acts across the datasources using a heuristic as shown in Table \ref{tab:norm_da}. Each of our models was evaluated on two thousand 8kHz anonymized in-house close-talk task-oriented audio conversations. The average number of turns in the audio dataset was 6 and was representative of the real world usage of task-oriented chatbots and followed the same conversation style and distribution as the text-only datasources mentioned above.

\begin{table}[t]
\small
\centering
\footnotesize
    \begin{tabular}{l|c}
    \toprule
    \textbf{Dialogue Act} & \textbf{Normalized DA}\\
    \midrule
    confirm, recommend, offerbook &  confirm\\
    inform, inform\_count & inform\\
    offer, offerbook, offer\_intent, select & offer\\
    request & request \\
    general-bye, goodbye & general-bye\\
    general-welcome & general-welcome\\
    general-reqmore & general-reqmore\\
    \bottomrule
    \end{tabular}
\caption{Normalization of system dialogue acts across data sources}
\label{tab:norm_da}
\vspace{-8mm}
\end{table}

\subsection{N-best rescoring setup}
The language model we used for first pass consisted of a standard 4-gram language model, trained on a weighted mix of out-of-domain and in-domain datasets. The weights were determined by minimizing the perplexity on an in-domain dev set. The Kneser-Ney (KN) \cite{KN:95} smoothed n-gram model estimated from the corpora had a final vocabulary of 500k words. All the neural language models had a word embedding size of 512. The BERT derived embeddings had a size of 768. In the LSTM LM models, we used a 3-layer LSTM, each of size 1,150. For transformer LMs, we used 6-layer Tranformer-XL decoder, each of size 512 with 4 attention heads. We used a segment and memory length of 15 while training. Both NLMs were trained with cross entropy objective loss function. During inference, we extract n-best hypothesis with \verb|n<=50| from the lattice generated by the first pass ASR model. We rescored the n-best hypothesis by multiplying the acoustic score with the acoustic scale and adding it to the scores obtained from the second pass NLM. 

\begin{table}[h]
\small
\centering
\footnotesize
    \begin{tabular}{l|c|c}
    \toprule
    \textbf{Model} & \textbf{PPLR} & \textbf{WERR}\\
    \midrule
    Non-contextual LSTM & - & -\\
    \midrule
    Context Carry Over &22.53\% & 1.65\%\\
    \hspace{0.5em} + BERT & 19.59\% & 4.58\%\\
    \midrule
    Feature Augmentation & & \\
    \hspace{0.5em} + Avg(BR) & 0.4\% & 2.06\%\\
    \hspace{0.5em} + Avg(BR;DA) & 9.61\% & 6.61\%\\
    \hspace{0.5em} + Attn(BR;DA) & 15.38\% & 2.89\%\\
    \hspace{0.5em} + Avg(BR;DA) + BERT & 2.67\% & 7.02\%\\
    \hspace{0.5em} + Attn(BR;DA) + BERT & 16.04\% & 5.78\%\\
    \bottomrule
    \end{tabular}
\caption{Relative Perplexity Reduction (PPLR) and the relative Word Error Rate Reduction (WERR) for the LSTM models. Attn: Attention, BR: Bot Response, DA: Dialogue Acts.}
\label{tab:lstm_report}
\vspace{-6mm}
\end{table}

\begin{table}[th]
\small
\centering
\footnotesize
    \begin{tabular}{l|c|c}
    \toprule
    \textbf{Model} & \textbf{PPLR} & \textbf{WERR}\\
    \midrule
    Non-contextual LSTM & - & -\\
    \midrule
    TXL & 9.98\% & 1.67\% \\
    \hspace{0.5em} + Context Carry Over (CCO) & 18.91\% & 3.34\%\\
    \midrule
    Masked Language Model (BERT) & & \\
    \hspace{0.5em} + Early Fusion (EF) & 31.63\% & 5.78\% \\
    \hspace{0.5em} + Cold Fusion (CF) & 32.93\% & 5.78\% \\
    \hspace{0.5em} + Simple Fusion (SF) & 37.53\% & 8.34\% \\
    \hspace{0.5em} + SF + CCO & 42.80\% & 9.16\%\\
    \bottomrule
    \end{tabular}
\caption{Relative Perplexity Reduction (PPLR) and the relative Word Error Rate Reduction (WERR) for the TXL models.}
\label{tab:txl_report}
\vspace{-6mm}
\end{table}

\section{Results and Discussion}
First, we compare the significance of incorporating auxiliary contextual signals and long context into recurrent (LSTM) and non recurrent (TXL) based NLMs. Tables \ref{tab:lstm_report} and \ref{tab:txl_report} summarize the overall relative perplexity reduction (PPLR) and the WERR for models with different settings. Then, we evaluate our best models on downstream NLU tasks and report the intent classification and slot labeling F1 scores in Table \ref{tab:e2e}. We also determine statistical significance of our WER improvements using matched pairs sentence segment word error test (MAPSSWE).

\vspace{2mm}
\noindent\textbf{Significance of System Dialogue Acts}: From our experiments in Table \ref{tab:lstm_report}, we observe that LSTM based LMs with CCO show high perplexity improvements, but WERR improvements are very marginal. We realize better WERR improvements by incorporating feature augmentation, which seems to help mitigate the problem of fuzzy distant context with LSTMs. Our best model was obtained by using system dialogue act as an additional feature. From the results, it is evident that dialogue acts, that are intended towards capturing the function of a dialogue turn, is an important contextual cue that the model learns to utilize effectively. 


\vspace{2mm}
\noindent\textbf{Effectiveness of BERT embeddings with Recurrent NLMs}: We conducted experiments to validate the extent of performance improvements that can be realized by using BERT embeddings extracted from user speech patterns. From results in Table \ref{tab:lstm_report},  it is clear that using BERT derived embeddings gives additional improvements in terms of WER when used in conjuction with CCO (1.65\% to 4.58\% ) and feature augmentation (average embeddings: 6.61\% to 7.02\%, attention based embeddings: 2.89\% to 5.78\%). Results show that the model is able to effectively attend to both the previous historical context and the rich semantic information condensed in the BERT embeddings generated from user provided speech patterns. Also, the observed perplexity gains indicate that the model is able to generate sharper probability distributions over domain specific vocabulary.

\vspace{2mm}
\noindent\textbf{Effectiveness of BERT embeddings with Non-Recurrent NLMs} : Table \ref{tab:txl_report} compares the improvements in PPL and WER, realized by incorporating BERT derived embeddings using different fusion techniques into a non-recurrent language model (vanilla TXL). In our experiments, we use TXL models with (memory=50) and without (memory=0) context. Our best performing model is obtained with SF + CCO setting, and has a relative improvement of 42.8\% and 9.16\% in terms of PPL, WERR respectively, over a strong LSTM baseline. This shows that fusion with BERT embeddings extracted from user provided speech patterns is an effective way to adapt NLMs to user specific domain. Also, from the qualitative results presented in Table \ref{tab:generated_text}, we can observe that text sampled from our best model (TXL + BERT SF) is salient to the user speech patterns.

\vspace{2mm}
\noindent\textbf{Performance on downstream NLU tasks}: To capture the impact on the end goals in these conversations, we measure the performance in terms of intent classification F1 score, slot labeling F1 score and content word error rate reduction (CWERR) on the ASR rescored outputs, using an NLU model. For computing CWERR, we remove all the stop words (commonly used function words, such as conjunctions and preposition) from the transcriptions and evaluate only on content words. From results in Table \ref{tab:e2e}, we can observe that our TXL models significantly outperform the LSTM models in terms of content word recognition. Though the LSTM LMs are able to improve on slot carrying auxiliary phrases, these gains fail to translate into actual slot recognition. The slot labeling F1 score improves by 0.48\% with the contextual LSTM LM and with a contetxual TXL this further improves significantly by 4.55\%.
 
 
 \begin{table}[t]
        \small
        \centering
        \footnotesize
        \begin{tabular}{l|c|c|c|c}
        \toprule
        \textbf{Model} & \textbf{CWERR} & \textbf{IC F1} & \textbf{SL F1} & \textbf{p-value}\\
        \midrule
        Non-contextual LSTM & & & & \\
        \hspace{0.5em}+ Avg(BR;DA) + BERT & 11.3\% & 0.33\% & 1.20\% & \textbf{0.031}\\
        \hspace{0.5em}+ Attn(BR;DA) + BERT & 6.08\% & 0.17\% & 0.48\% & 0.195\\
        \hspace{0.5em}+ CCO + BERT & 6.08\% & 0.17\% & 0.48\% & 0.052\\
        \midrule
        TXL & 12.60\% & 0.46\% & 3.36\% & \textbf{0.026}\\
        \hspace{0.5em}+ CCO & 15.65\% & 0.46\% & 3.57\% & \textbf{0.003}\\
        \hspace{0.5em}+ BERT SF & 18.26\% & 1.05\% & 4.15\% & \textbf{0.002}\\
        \hspace{0.5em}+ BERT SF + CCO & 18.26\% & 1.20\% & 4.55\% & \textbf{0.004}\\
        \bottomrule
        \end{tabular}
        \caption{Relative reduction in content WER (CWERR), intent classification (IC) F1 and slot labeling (SL) F1 and MAPSSWE p-value test on WER with significant improvements in bold.}
        \label{tab:e2e}
    \vspace{-6mm}
    \end{table}
    
\begin{table}[t]
    \begin{tabular}{p{8cm}}
    \toprule
    \footnotesize
    \textbf{Input Prompt:} "<sos> can you"  \\
    \footnotesize
    \textbf{BERT Embed:} None  \\
    \footnotesize
    \textbf{Generated Text:} "send a copy of the details please <eos>"  \\
    \midrule
    \footnotesize
    \textbf{Input Prompt:} "<sos> can you"  \\
    \footnotesize
    \textbf{BERT Embed:} "travel"  \\
    \footnotesize
    \textbf{Generated Text:} "tell me what amenities are offered at the hotel <eos>"  \\
    \midrule
    \footnotesize
    \textbf{Input Prompt:} "<sos> can you"  \\
    \footnotesize
    \textbf{BERT Embed:} "bank"  \\
    \footnotesize
    \textbf{Generated Text:} "show me all the transfers scheduled <eos>"  \\
    \bottomrule
    \end{tabular}
    \caption{Example text sampled from TXL + BERT SF model. The model is primed with an input prompt and a domain specific BERT embedding. When compared to providing an empty BERT embedding, the model adapts to the speech patterns when a domain specific BERT embedding is provided.}
    \label{tab:generated_text}
\vspace{-6mm}
\end{table}

\section{Conclusions}

In this paper we explored different ways to incorporate auxiliary context information along with cross utterance context to improve LSTM and Transformer-XL based NLMs to rescore n-best list from a hybrid ASR system. We introduced a new method that successfully adapts both recurrent and non-recurrent based NLMs towards user provided speech patterns on-the-fly by using semantic pretrained BERT embeddings. Additionally, we show that NLM rescoring in task oriented dialogue systems can be improved significantly by combining context carry over with feature based augmentation that includes system dialogue acts. Experiments on task oriented audio conversations show substantial WER and PPL gains which also carry over to downstream NLU metrics such as improved intent classification and slot labeling F1 scores. Future work can look at generalization of the improvements on out of domain utterances and utilizing these methods to rescore hypothesis from an end-to-end ASR system.

\bibliographystyle{IEEEtran}

\bibliography{main}

\end{document}